# Logical Inference Algorithms and Matrix Representations for Probabilistic Conditional Independence

**Mathias Niepert**
Department of Computer Science
Indiana University
Bloomington, IN, USA
mniepert@indiana.edu

## Abstract

Logical inference algorithms for conditional independence (CI) statements have important applications from testing consistency during knowledge elicitation to constraint-based structure learning of graphical models. We prove that the implication problem for CI statements is decidable, given that the size of the domains of the random variables is known and fixed. We will present an approximate logical inference algorithm which combines a falsification and a novel validation algorithm. The validation algorithm represents each set of CI statements as a sparse 0-1 matrix **A** and validates instances of the implication problem by solving specific linear programs with constraint matrix **A**. We will show experimentally that the algorithm is both effective and efficient in validating and falsifying instances of the probabilistic CI implication problem.

## 1 Introduction

Knowledge elicitation is an important task in the field of reasoning under uncertainty [1]. For example, consider the problem of eliciting knowledge from several domain experts in an attempt to model a probabilistic system (e.g., a Bayesian or Markov network). The resulting incomplete expert feedback might be a combination of some specific subjective probabilities, (conditional) independence and dependence information for the random variables under consideration, and conditional probabilities. Furthermore, in some cases, statistical tests on different heterogeneous data sets could provide additional sources of evidence. Each of these bits of information can be interpreted as a *constraint* on the joint probability distribution one wants to model. Finding a suitable model can then be interpreted as a *constraint satisfaction problem* (CSP), and the approach to harness CSP solvers for instances of this and related problems has been known for more than 10 years (Druzdzel and van der Gaag [2], Dechter [3]). However, (conditional) independence and dependence statements pose a special problem, because they often introduce *non-linear* constraints which mostly result in infeasible CSP instances. Therefore, a remaining important challenge in the context of knowledge elicitation is to test for *consistency* of the (conditional) independence and dependence information that has been collected from different sources. For this to be possible, one would need an algorithm that decides the implication problem for CI statements (Geiger and Pearl [4]), that is, an algorithm that can infer CI statements which are logically implied by a set of given ones, relative to the class of discrete probability measures. There are several other important applications of an inference algorithm for CI statements. For example, the representation of CI information is mostly based on the well-known semi-graphoid axioms of independence. There are ways to improve on this representation by using the notion of o-dominant triplets (Studený [5]) and stable independence (de Waal and van der Gaag [6]), both of which need fast logical inference algorithms. Furthermore, inference algorithms for CI statements have been used to learn the structure of graphical models using only few independence tests (Gandhi et al. [7]).

In this paper, we will first prove that the implication problem for CI statements relative to the class of discrete (positive) probability measures is decidable, given that the cardinalities of the domains of all random variables are known and fixed. We achieve this by encoding instances of the implication problem as sentences in the first-order theory over the reals with addition and multiplication. An instance of the implication problem holds if and only if the corresponding sentence in the first-order theory has no model. The decidability then follows since the first-order theory over the reals with addition and multiplication is de-cidable (Tarski [8]). Since encoding instances of the implication problem as decision problems in the first-order theory over the reals is not feasible for practical applications, we will introduce an *approximate logical inference* algorithm which combines a *falsification* and a novel *validation* algorithm. The validation algorithm represents each set of CI statements as a sparse 0-1 matrix $\mathbf{A}$, and validates instances of the implication problem by solving linear programs with constraint matrix $\mathbf{A}$. Thus, by only requiring the algorithm to decide most but not all instances of the implication problem, we can leverage linear constraint solvers for our purposes. In an extensive experimental section, we will demonstrate this inference algorithm to be both effective and efficient in validating and falsifying instances of the implication problem.

## 2 Preliminaries

**Definition 2.1.** A *probability model* over $S = \{s_1, \dots, s_n\}$ is a pair $(dom, P)$, where $dom$ is a domain mapping that maps each $s_i$ to a finite domain $dom(s_i)$, and $P$ is a probability measure having $dom(s_1) \times \cdots \times dom(s_n)$ as its sample space. For $A = \{a_1, \dots, a_k\} \subseteq S$, we will say that $\mathbf{a}$ is a domain vector of $A$ if $\mathbf{a} \in dom(a_1) \times \cdots \times dom(a_k)$.

In what follows, we will only refer to probability measures, keeping their probability models implicit.

**Definition 2.2.** Let $I(A, B|C)$ be a CI statement, and let $P$ be a probability measure. We say that $P$ *satisfies* $I(A, B|C)$, and write $\models_P I(A, B|C)$, if for every domain vector $\mathbf{a}$, $\mathbf{b}$, and $\mathbf{c}$ of $A$, $B$, and $C$, respectively, $P^C(\mathbf{c})P^{ABC}(\mathbf{a}, \mathbf{b}, \mathbf{c}) = P^{AC}(\mathbf{a}, \mathbf{c})P^{BC}(\mathbf{b}, \mathbf{c})$.

## 3 On the Decidability of Implication Problems for CI Statements

In this section we will investigate the decidability of the implication problem for CI statements relative to the class of discrete probability measures. One of the key ideas is to leverage the fact that every discrete probability measure is fully characterized by specifying a finite number of probability densities. For an in-depth discussion of related observations and ideas in the context of the probability calculus we refer the reader to (Fitelson [19]). Using this idea, we will be able to show that the implication problem for probabilistic conditional independence statements can be encoded as a sentence in the first-order theory over the reals with addition and multiplication, given that the cardinalities of the domains of the random variables are fixed a priori, which is an assumption often made for probabilistic models. This will, for instance, imply the decidability of the implication problem for

$$\Phi = \exists d_1 ... \exists d_N ($$

$$\bigwedge_{I(A,B|C)\in\mathcal{C}} \Bigg( \bigwedge_{\mathbf{s}\in dom(s_1)\times\cdots\times dom(s_n)} \big(\varphi(P^{ABC}(\mathbf{s}|_{ABC}))\cdot\varphi(P^{C}(\mathbf{s}|_{C})) = \varphi(P^{AC}(\mathbf{s}|_{AC}))\cdot\varphi(P^{BC}(\mathbf{s}|_{BC}))\big)\Bigg) \quad (1)$$

$$\wedge \bigwedge_{I(A,B|C)\in\mathcal{D}} \Bigg( \bigvee_{\mathbf{s}\in dom(s_1)\times\cdots\times dom(s_n)} \big(\varphi(P^{ABC}(\mathbf{s}|_{ABC}))\cdot\varphi(P^{C}(\mathbf{s}|_{C})) \neq \varphi(P^{AC}(\mathbf{s}|_{AC}))\cdot\varphi(P^{BC}(\mathbf{s}|_{BC}))\big)\Bigg) \quad (2)$$

$$\wedge \bigwedge_{\mathbf{s}\in dom(s_1)\times\cdots\times dom(s_n)} \varphi(P(\mathbf{s})) \geq 0 \quad (3)$$

$$\wedge \quad d_1 + ... + d_N = 1) \quad (4)$$

Figure 1: First-order sentence encoding instances of the implication problem for conditional independence statements. Here we write $(a \neq b)$ for $\neg(a = b)$.

CI statements relative to the class of binary discrete probability measures. Let us first define the implication problem relative to the notion of *satisfaction* from Definition 2.2.

**Definition 3.1.** Let $S$ be a finite set of random variables, let $\mathcal{C}$ and $\mathcal{D}$ be two non-empty sets of CI statements over $S$, let $\mathcal{P}$ be the class of discrete probability measures over $S$. We say that $\mathcal{C}$ *implies* $\mathcal{D}$ relative to $\mathcal{P}$, and write $\mathcal{C} \models \mathcal{D}$, if every $P \in \mathcal{P}$ that *satisfies* each CI statement in $\mathcal{C}$ also *satisfies* at least one CI statement in $\mathcal{D}$. We will denote the corresponding decision problem with $(S, \mathcal{C}, \mathcal{D})$.

This is the *most general* definition of the decision problem which is necessary for cases where perfect models are not guaranteed. For a discussion of perfect models and related logical and algorithmic concepts of probabilistic conditional independence, we refer the reader to (Geiger and Pearl [4]). We can now prove the decidability of the implication problem for conditional independence statements under the additional assumption that the cardinalities of the domains of all random variables are known and fixed.

**Theorem 3.2.** *Let $k_1, ..., k_n \in \mathbf{N}$, let $S = \{s_1, ..., s_n\}$ be a set of $n$ discrete random variables with $|s_i| = k_i$ for $1 \leq i \leq n$, and let $\mathcal{C}$ and $\mathcal{D}$ be sets of CI statements over $S$. Then $(S, \mathcal{C}, \mathcal{D})$ is decidable.*

*Proof.* Let $N = \prod_{s_i \in S} |s_i| = \prod_{i=1}^{n} k_i$. Let $\mathcal{C}$ and $\mathcal{D}$ be two sets of CI statements over $S$. Every discrete probability measure $P$ over $S$ can be fully characterized by $N$ probability densities $P(\mathbf{s})$, for all $\mathbf{s} \in dom(s_1) \times \cdots \times dom(s_n)$. Every (marginal) probabil-ity can then be expressed as a sum of densities: For every $A = \{a_{i_1}, \ldots, a_{i_k}\} \subseteq S$ and $\mathbf{a} \in dom(a_{i_1}) \times \cdots \times dom(a_{i_k})$ we have that $P^A(\mathbf{a}) = \sum_{\{\mathbf{s} \text{ with } \mathbf{s}|A=\mathbf{a}\}} P(\mathbf{s})$ where $\mathbf{s}|A$ is the projection of $\mathbf{s}$ onto $A$. Let $\lambda : dom(s_1) \times \cdots \times dom(s_n) \rightarrow \{1, ..., N\}$ be a bijection that maps each element in $dom(s_1) \times \cdots \times dom(s_n)$ to a number in $\{1, ..., N\}$. Let $\varphi$ be the function that maps each marginal probability to the *formula* in the first-order theory over the reals that *encodes* the corresponding sum of density symbols indexed according to $\lambda$. We can now encode the instance of the implication problem as a sentence in the first-order theory with addition and multiplication as depicted in Figure 1. Tarski showed that the first-order theory over the reals with addition and multiplication is decidable, by providing an algorithm (using quantifier elimination) that can return, on any input sentence $\Psi$ and in a finite number of steps, whether or not there exists a model $\mathbf{M}$ of $\Psi$ [8]. Hence, what remains to be shown is that there exists a model $\mathbf{M}$ of $\Phi$ if and only if $\mathcal{C} \not\models \mathcal{D}$. Let $P$ be the measure that corresponds to $\mathbf{M}$. Since $\mathbf{M}$ satisfies (3) and (4), $P$ satisfies the first and the second of Kolmogorov's axioms. Furthermore, by definition of $\mathbf{M}$ and $P$, $P$ satisfies the third of Kolmogorov's axioms. Hence, $\mathbf{M}$ corresponds to a discrete probability measure over $S$. Since $\mathbf{M}$ satisfies (1) and (2) we have that $P$ satisfies every $I(A, B|C) \in \mathcal{C}$ but no $I(A, B|C) \in \mathcal{D}$. This implies $\mathcal{C} \not\models \mathcal{D}$. Finally, assume that $\mathcal{C} \not\models \mathcal{D}$. Hence, there exists a discrete probability measure $P$ over $S$ such that $P$ satisfies every $I(A, B|C) \in \mathcal{C}$ but no $I(A, B|C) \in \mathcal{D}$. Let $\mathbf{M} \in \mathbb{R}^N$ and let the $i$-th element of $\mathbf{M}$ be the density $P(\lambda^{-1}(i))$. It follows that $\mathbf{M}$ is a model of $\Phi$. □

Of course, the theorem implies the decidability of the implication problem relative to the class of binary discrete probability measures. Note that the size of $\Phi$ (that is, the number of polynomials) grows exponentially with the number of variables. Indeed, each of the first-order sentences $\Phi$ over binary variables will contain an exponential number of polynomials since each CI statement in $\mathcal{C}$ leads to $2^{|S|}$ conjunctions of different polynomials in $\Phi$. By replacing the inequalities $\geq$ in part (3) of the first-order sentence $\Phi$ with strict inequalities $>$, we can derive the following corollary.

**Corollary 3.3.** *Let $k_1, ..., k_n \in \mathbf{N}$, let $S = \{s_1, ..., s_n\}$ be a set of $n$ discrete random variables with $|s_i| = k_i$ for $1 \leq i \leq n$, and let $\mathcal{C}$ and $\mathcal{D}$ be sets of CI statements over $S$. Then $\mathcal{C} \models \mathcal{D}$ is decidable relative to the class of positive discrete probability measures over $S$.*

Even though considerable progress has been made in improving Tarski's quantifier elimination algorithm (for a survey of methods we refer the reader to Collins [10]), the method remains infeasible even for small instances. The time complexity of deciding whether a model for a first-order sentence with addition and multiplication exists over the reals is doubly-exponential in the worst case, and in the case of instances of the CI implication problem, the input is already exponential in the size of $S$. However, there are ways to *approximate* the decision of the implication problem by combining powerful *falsification* and *validation* algorithms (Bouckaert and Studený [9]). These algorithms falsify and validate instances of the implication problem relative to the class of discrete probability measures with *arbitrary* finite domains, which makes them also applicable to instances of the implication problem where the domain cardinalities are fixed.

## 4 Approximate Logical Inference

We will harness one of the recently discovered falsification algorithms and introduce a novel validation algorithm that encodes instances of the implication problem as linear programs with sparse 0-1 constraint matrices. Our goal is to compute, for any given set of CI statements $\mathcal{C}$ over a finite set $S$, those CI statements over $S$ that are implied by $\mathcal{C}$ and those that are not implied. In order to achieve this, we will first apply the falsification algorithm to falsify a large fraction of invalid instances of the implication problem and also use it to compute the input to the validation algorithm. Furthermore, we will be able to show that the validation algorithm is closely related to the theory of structural and combinatorial imsets (Studený [17]). Before we discuss the technical details of the validation algorithm, let us first recall the falsification algorithm introduced in (Niepert et al. [12]).

### 4.1 Falsification Algorithm

Since our objective is the implication problem relative to the class of discrete probability measures with arbitrary finite domains, for which perfect models exist (Geiger and Pearl [4]), deciding $\mathcal{C} \models \mathcal{D}$ is equivalent to deciding whether $\mathcal{C} \models d$ for at least one $d \in \mathcal{D}$. Therefore, in the remainder of this section, we will focus on algorithms that decide the implication problem $\mathcal{C} \models c$ for a set of CI statements $\mathcal{C}$ and a single CI statement $c$. This will simplify both the notation and the technical parts of the following sections.

Given two subsets $A$ and $B$ of $S$, we will write $[A, B]$ for the lattice $\{U \mid A \subseteq U \ \& \ U \subseteq B\}$. Using the notion of a lattice, we can associate semi-lattices with conditional independence statements.

**Definition 4.1.** Let $I(A, B|C)$ be a CI statement. The *semi-lattice* of $I(A, B|C)$ is defined by $\mathcal{L}(A, B|C) = [C, S] - ([A, S] \cup [B, S])$.

We will write $\mathcal{L}(c)$ to denote the semi-lattice of a conditional independence statement $c$, and $\mathcal{L}(\mathcal{C})$ to denote the union of semi-lattices, $\bigcup_{c' \in \mathcal{C}} \mathcal{L}(c')$, of a set of CI statements $\mathcal{C}$. These semi-lattices can be used to falsify instances of the implication problem.

**Proposition 4.2** (Niepert et al. [12])**.** *Let $\mathcal{C}$ be a set of CI statements and $c$ be a single CI statement. If $\mathcal{L}(\mathcal{C}) \not\supseteq \mathcal{L}(c)$, then $\mathcal{C} \not\models c$.*

**Example 4.3.** Let $S = \{a, b, c, d\}$, let $\mathcal{C} = \{I(a, b|cd), I(a, d|bc)\}$ and let $I(a, bd|c)$ be a single CI statement. Then, $\mathcal{L}(\mathcal{C}) = \mathcal{L}(a, b|cd) \cup \mathcal{L}(a, d|bc) = \{cd\} \cup \{bc\} = \{cd, bc\}$ and $\mathcal{L}(a, bd|c) = \{c, bc, cd\}$. Since $\mathcal{L}(\mathcal{C}) \not\supseteq \mathcal{L}(a, bd|c)$ we have that $\mathcal{C} \not\models I(a, bd|c)$.

It has been shown that testing for semi-lattice inclusion is a coNP-complete decision problem (Niepert and Van Gucht [13]). However, we also know that there exists a linear time reduction to SAT and that we can leverage SAT solvers to decide semi-lattice inclusion very efficiently (without storing the exponentially sized semi-lattices), even for instances of up to several hundreds of variables (Niepert and Van Gucht [13]). Indeed, we discovered in our experiments that the critical and most time-consuming part of the approximate logical inference algorithm is not the falsification but the validation algorithm.

### 4.2 Validation Algorithm

In general, a validation algorithm takes as input an instance of the implication problem, consisting of a set of CI statements $\mathcal{C}$ and a single CI statement $c$ over a finite set $S$, and accepts only if $\mathcal{C} \models c$. Of course, the algorithm not accepting an instance of the implication problem does not imply that the instance is invalid. Please note that one of the most prominent validation algorithms is the algorithm that computes the closure of the semi-graphoid axioms (Dawid [11], Pearl [16]). However, the closure of the semi-graphoid axioms can only validate a small fraction of the set of verifiable instances. Before we can derive our validation algorithm, we need some definitions of important technical concepts.

**Definition 4.4.** Let $P$ and $Q$ be two probability measures over a discrete sample space, and let $P$ be absolutely continuous with respect to $Q$. Then, the relative entropy (Kullback-Leibler divergence) $H$ is defined as

$$H(P|Q) := \sum_{\mathbf{x}} \{P(\mathbf{x}) \log \frac{P(\mathbf{x})}{Q(\mathbf{x})},\ P(\mathbf{x}) > 0\},$$

with $\mathbf{x}$ ranging over all elements of the discrete sample space.

**Definition 4.5** (Studený [17])**.** Let $P$ be a probability measure, and let $H$ be the relative entropy. The *multi-information function* $M_P : 2^S \to [0, \infty]$ induced by $P$ is defined as

$$M_P(A) := H(P^A | \prod_{a \in A} P^{\{a\}}),$$

for each non-empty subset $A$ of $S$ and $M_P(\emptyset) = 0$.

**Definition 4.6.** Let $S$ be a finite set, and let $F$ be a real-valued function over $S$. The *Möbius inversion* of $F$ is the real-valued function $\Delta F$ defined by $\Delta F(X) = \sum_{X \subseteq U \subseteq S} (-1)^{|U|-|X|} F(U)$, for each $X \subseteq S$.

Now we have the following crucial relationship between a multiinformation function, its Möbius inversion, and the semi-lattice of a CI statement.

**Lemma 4.7.** *Let $S$ be a finite set of random variables, let $P$ be a discrete probability measure over $S$, let $M_P$ be the multiinformation function induced by $P$, let $\Delta M_P$ be the Möbius inversion of $M_P$, and let $I(A, B|C)$ be a CI statement over $S$. Then, the following statements are equivalent*

*(1)* $P$ *satisfies* $I(A, B|C)$

*(2)* $M_P(ABC) + M_P(C) - M_P(AC) - M_P(BC) = 0$

*(3)* $\sum_{U \in \mathcal{L}(A,B|C)} \Delta M_P(U) = 0$

*Proof.* Studený showed that (1) if and only if (2) [17]. In addition, we have that (2) if and only if (3), because $F(ABC) + F(C) - F(AC) - F(BC) = \sum_{U \in \mathcal{L}(A,B|C)} \Delta F(U)$ for any real-valued function $F$. We refer the reader to Sayrafi and Van Gucht [14] for a proof of the later statement. □

We will now be able to harness the equivalences stated in the previous lemma to represent each set of CI statements $\mathcal{C}$ as a *minimal* sparse 0-1 matrix $\mathbf{A}$. Each instance of the implication problem with $\mathcal{C}$ as the set of antecedents will then correspond to a linear program with equality constraints specified by $\mathbf{A}$. Before we explain the construction of the constraint matrix $\mathbf{A}$, however, we have to define some additional technical concepts. For some of the following results, we need the concept of elementary CI statements, which are simply CI statements $I(a, b|K)$ with $a, b \in S$ and $K \subseteq S \setminus \{a, b\}$. We will write $\mathcal{B}(S)$ to denote the set of elementary CI statements over a finite set $S$.

**Definition 4.8.** Let $S$ be a finite set and let $\mathcal{C}$ be a set of CI statements over $S$. The set of *relevant* elementary CI statements $\mathcal{R}(\mathcal{C})$ is defined as follows:

$$\mathcal{R}(\mathcal{C}) = \{I(a, b|K) \in \mathcal{B}(S) \mid \mathcal{L}(a, b|K) \subseteq \mathcal{L}(\mathcal{C})\}.$$

We will use the elementary CI statements in $\mathcal{R}(\mathcal{C})$ to construct the constraint matrix $\mathbf{A}$. Please note that $\mathcal{R}(\mathcal{C})$ is the set of elementary CI statements over $S$ that remain after the application of the falsification algorithm (Niepert et al. [12]). Hence, it follows that polynomial-time heuristics and SAT solvers can be employed to compute the set $\mathcal{R}(\mathcal{C})$ efficiently for up to several hundreds of variables (Niepert and Van Gucht [13]). By Proposition 4.2, only CI statements $I(A,B|C)$ with $\mathcal{L}(A,B|C) \subseteq \mathcal{L}(\mathcal{C})$ can possibly be implied by $\mathcal{C}$. Now, using the concept of a semi-lattice, each of these candidate CI statements $c = I(A,B|C)$ can be written as a vector $\mathbf{v}_c$ relative to the space $\{0,1\}^{\mathcal{L}(\mathcal{C})}$ as follows: For every $U \in \mathcal{L}(\mathcal{C})$ we have $\mathbf{v}_c(U) = 1$ if $U \in \mathcal{L}(A,B|C)$ and $\mathbf{v}_c(U) = 0$ otherwise. The vector representation of a set of CI statements $\mathcal{C}$ can then be defined as the sum of the vectors corresponding to individual elements in $\mathcal{C}$: $\mathbf{v}_{\mathcal{C}} = \sum_{c\in\mathcal{C}} \mathbf{v}_c$. This is equivalent to the definition of an imset (Studený [17]), except that we use the Möbius inversion to avoid negative elements in the vector representation and that the vector representation is constructed relative to the union of semi-lattices $\mathcal{L}(\mathcal{C})$ of the CI statements in $\mathcal{C}$. Given these definitions of vector representations for individual CI statements and for sets of CI statements, we can state the following crucial result.

**Proposition 4.9.** *Let $S$ be a finite set, let $\mathcal{C}$ be a set of CI statements, let $c$ be a single CI statement over $S$ and let $\mathbb{Q}^+$ be the non-negative rational numbers. Then, $\mathcal{C} \models c$ if*

$$\mathbf{v}_{\mathcal{C}} = \mathbf{v}_c + \textstyle\sum_{r\in\mathcal{R}(\mathcal{C})} k_r \cdot \mathbf{v}_r \text{ with } k_r \in \mathbb{Q}^+. \quad (5)$$

*Proof.* Let $P$ be a probability measure that satisfies all CI statements in $\mathcal{C}$ and let $\Delta M_P$ be the Möbius inversion of the multiinformation function $M_P$ induced by $P$. Since $M_P$ is a supermodular function (Studený [17]), we have $\sum_{r\in\mathcal{R}(\mathcal{C})} \cdot (k_r \cdot \sum_{U\in\mathcal{L}(r)} \Delta M_P(U)) \geq 0$, and also $\sum_{U\in\mathcal{L}(c)} \Delta M_P(U) \geq 0$. Now, since $P$ satisfies all CI statements in $\mathcal{C}$ we have that $\sum_{c'\in\mathcal{C}} \sum_{U\in\mathcal{L}(c')} \Delta M_P(U) = 0$ by Lemma 4.7. Since equality (5) holds by assumption, we have that $\sum_{c'\in\mathcal{C}} \sum_{U\in\mathcal{L}(c')} \Delta M_P(U) = \sum_{U\in\mathcal{L}(c)} \Delta M_P(U) + \sum_{r\in\mathcal{R}(\mathcal{C})} \cdot (k_r \cdot \sum_{U\in\mathcal{L}(r)} \Delta M_P(U)) = 0$. Hence, $\sum_{U\in\mathcal{L}(c)} \Delta M_P(U) = 0$ and by Lemma 4.7 it follows that $P$ satisfies $c$. □

In light of these results, we can now rewrite equation (5) in the previous proposition as a linear program (Schrijver [18]). A linear program has the form

$$\text{minimize } \mathbf{c}^T\mathbf{x} \quad (6)$$

$$\text{subject to } \mathbf{A}\mathbf{x} \text{ eq } \mathbf{b}, \mathbf{x} \geq 0 \quad (7)$$

where eq is one of $\{\leq, \geq, =\}$. For our purposes, eq is the equality sign, the columns of matrix $\mathbf{A}$ are the vectors $\mathbf{v}_r$ for each of the relevant elementary CI statements, that is, for the CI statements in $\mathcal{R}(\mathcal{C})$, and $\mathbf{b} = \mathbf{v}_{\mathcal{C}} - \mathbf{v}_c$. Clearly, our objective function is the zero-function $\mathbf{0}^T$ because we are only interested in the existence of a solution for the equality constraints. This is often referred to as the *feasibility problem* of finding a solution for the system of linear constraints.

**Example 4.10.** Let $S = \{a,b,c,d\}$ and let $\mathcal{C} = \{I(a,b|\emptyset), I(c,d|a), I(c,d|b), I(a,b|cd)\}$. Then, $\mathcal{R}(\mathcal{C}) = \{I(a,b|\emptyset), I(a,b|c), I(a,b|d), I(a,b|cd), I(c,d|\emptyset), I(c,d|a), I(c,d|b), I(c,d|ab)\}$ and $\mathcal{L}(\mathcal{C}) = \{\emptyset, a, b, c, d, ab, cd\}$. The columns of the *minimal* 0-1 matrix $\mathbf{A}$ below correspond to the eight *relevant* elementary CI statements and the number of rows is determined by $\mathcal{L}(\mathcal{C})$.

$\mathbf{A} =$

| | $\mathbf{e}_1$ | $\mathbf{e}_2$ | $\mathbf{e}_3$ | $\mathbf{e}_4$ | $\mathbf{e}_5$ | $\mathbf{e}_6$ | $\mathbf{e}_7$ | $\mathbf{e}_8$ |
|---|---|---|---|---|---|---|---|---|
| cd | 1 | 1 | 1 | 1 | 0 | 0 | 0 | 0 |
| ab | 0 | 0 | 0 | 0 | 1 | 1 | 1 | 1 |
| a | 0 | 0 | 0 | 0 | 1 | 1 | 0 | 0 |
| b | 0 | 0 | 0 | 0 | 1 | 0 | 1 | 0 |
| c | 1 | 1 | 0 | 0 | 0 | 0 | 0 | 0 |
| d | 1 | 0 | 1 | 0 | 0 | 0 | 0 | 0 |
| ∅ | 1 | 0 | 0 | 0 | 1 | 0 | 0 | 0 |

We have that $\mathbf{v}_{\mathcal{C}}^T = (2,2,1,1,1,1,1)$. Now, let $I(c,d|\emptyset)$ be a CI statement. Then we have that $\mathbf{b}^T = \mathbf{v}_{\mathcal{C}}^T - \mathbf{v}_{I(c,d|\emptyset)}^T = (2,2,1,1,1,1,1) - (0,1,1,1,0,0,1) = (2,1,0,0,1,1,0)$. Finally, it follows that $\mathcal{C} \models I(c,d|\emptyset)$ since $\mathbf{b} = \mathbf{e}_2 + \mathbf{e}_3 + \mathbf{e}_8$.

It is well-known that linear programs (LPs) are solvable in polynomial time in the number of variables. However, in the worst case the reduction leads to an LP with an exponential number of variables ($\binom{|S|}{2}2^{|S|-2}$; the maximum number of elementary CI statements over $S$) and constraints ($2^{|S|} - |S| - 1$; the maximum cardinality of the set $\mathcal{L}(\mathcal{C})$). As a rule of thumb, the more columns matrix $\mathbf{A}$ has the more difficult is the corresponding LP problem. An advantage of our method over a naïve approach is that $\mathbf{A}$ only consists of the vectors representing the *relevant* elementary CI statements $\mathcal{R}(\mathcal{C})$. This means that the number of columns (that is, the number of variables of the LP) can be very small compared to the worst case. In rare cases, the solutions to the LPs might be inaccurate due to round-off and truncation errors. Therefore, when we obtain a solution, we expand the elements of the solution vector into fractions of integers, which results in a vector $\mathbf{x}_f$, and only accept if $\mathbf{A}\mathbf{x}_f = \mathbf{b}$. We also would like to underscore that matrix $\mathbf{A}$ is always a 0-1 matrix, leading to better numerical stability and the possibility to employ existing sparse matrix data structures. We will come back to algorithmic issues when we discuss the results of our experiments.

### 4.3 Combinatorial and Structural Imsets

There is a close link to Studený's theory of imsets [17], on which we will briefly elaborate in this section. Let $\mathcal{C}$ be a set of CI statements and let $c$ be a CI statement over a set $S$. Then, under the assumption that we can ignore numerical inaccuracies, one can test whether imset $u_{\mathcal{C}} - u_c$ is structural using the previously introduced reduction to a linear program. Furthermore, one can test whether the imset is combinatorial by reducing it to the identical integer program. Again, we want to stress that numerical rounding and truncation errors might lead to inaccurate results, and, therefore, the method should be used with caution when mathematical properties about combinatorial and structural imsets are to be proved. However, one of the results of our experiments is that the solver of the LP instances delivered integer and small rational solutions in all but some cases which allowed us to verify their correctness. We refer the reader to Hemmecke et al. [20] who used, among other tools, integer programming to find a structural imset which is not combinatorial.

## 5 Experiments

We will try to mainly address the following empirical questions with our experiments:

1. **Effectiveness**: What fraction of the instances of the implication problem can we either falsify or validate?
2. **Efficiency**: How fast does the algorithm run; to how many variables does it scale? How much more efficient is the algorithm compared to the naïve approach both in terms of time and space complexity?
3. **Structural and Numerical properties**: How large is the constraint matrix **A** for different instances? What are the numerical properties of the solutions?

To judge the effectiveness and efficiency of the algorithm we must apply it to instances of the implication problem over different number of variables. Since the distribution of implication problems in real-world applications is unknown, our experiments need to be run on randomly generated instances. Using the method of randomly generating test instances from (Bouckaert and Studený [9]) allows us to compare the experimental outcomes with existing results. Hence, for each experiment we first generated instances of the implication problem $(S, \mathcal{C}, c)$ by randomly selecting $n$ different sets of elementary CI statements over $S$ as antecedents $\mathcal{C}$, and for each of these, $k$ different elementary CI statements $c$ over $S$ as consequence, one at a time. We first applied the falsification algorithm to these instances and used it to create the constraint matrix **A** and vector **b** from $\mathcal{C}$ and $c$ as described in the previous section. For the resulting linear programs we used lp_solve[1] an open-source linear programming system that can solve both linear and integer programs. It is based on the revised simplex method and the branch-and-bound method for integer programs. We did not change the standard optimization settings of the solver. Furthermore, we only accepted a solution if its rational expansion solved the respective constraints. For our purposes this is unproblematic because the objective is to validate as many instances of the implication problem as possible while entirely ruling out false positives. All experiments were run on a dual-core 3.2GHz Linux PC with 2GB RAM.

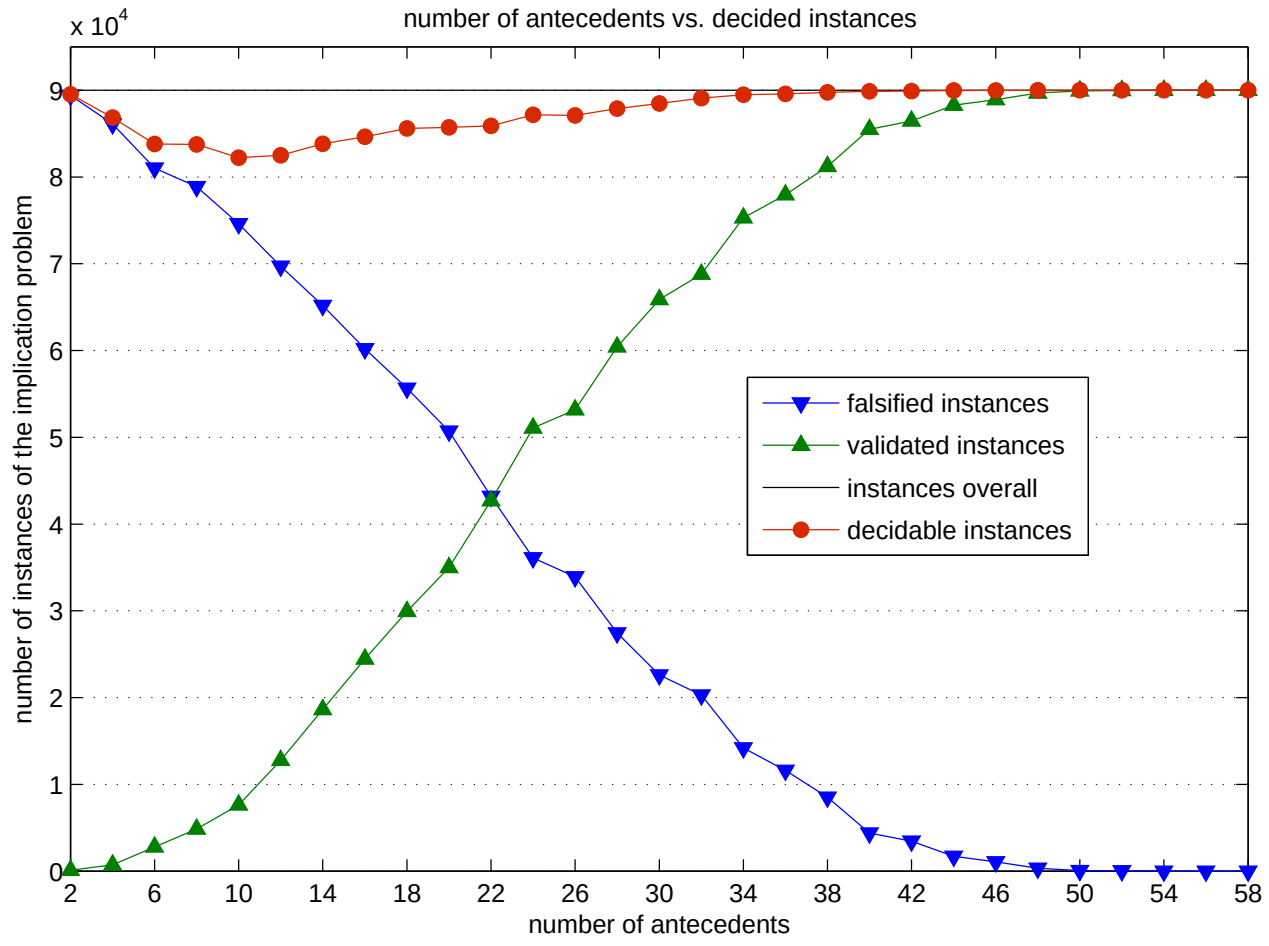

Figure 2: Falsification and validation curves of the approximate decision algorithm for five variables. The curve with circular markers depicts the number of instances that could either be falsified or validated.

Figure 2 shows the number of instances that could either be validated or falsified by the algorithms for five variables. For each $\ell = 2, ..., 58$ (the number of antecedents) we randomly created 4,500 different sets of $\ell$ elementary CI statements, and for each of those randomly selected 20 different elementary CI statements as consequences, one at a time, resulting in 90,000 instances of the implication problems for each $\ell$. The results show that only a small fraction of the instances could not be decided and that for larger values of $\ell$ (for five variables: $\ell > 40$) all of the instances could either be falsified or validated. This behavior of the algorithm was consistent over all tested number of variables (4,...,15).

[1]Michel Berkelaar, Kjell Eikland, and Peter Notebaert. lp_solve, an open source (Mixed-Integer) Linear Programming system originally developed at Eindhoven University of Technology; http://lpsolve.sourceforge.net/

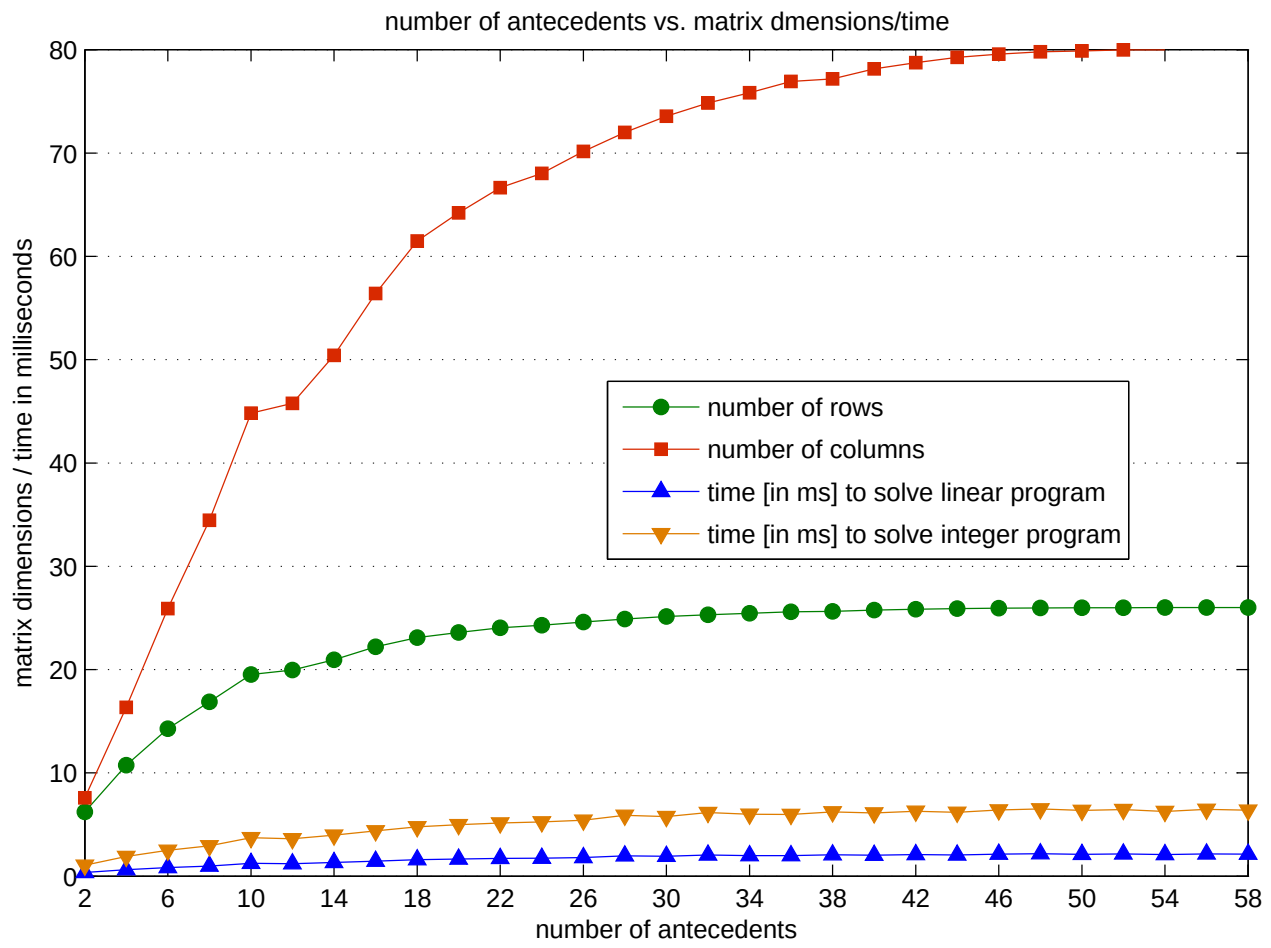

Figure 3: Dimensions of matrix $\mathbf{A}$ that encodes a set of $x$ antecedents over five variables; and time in milliseconds needed to solve the corresponding linear and integer programs.

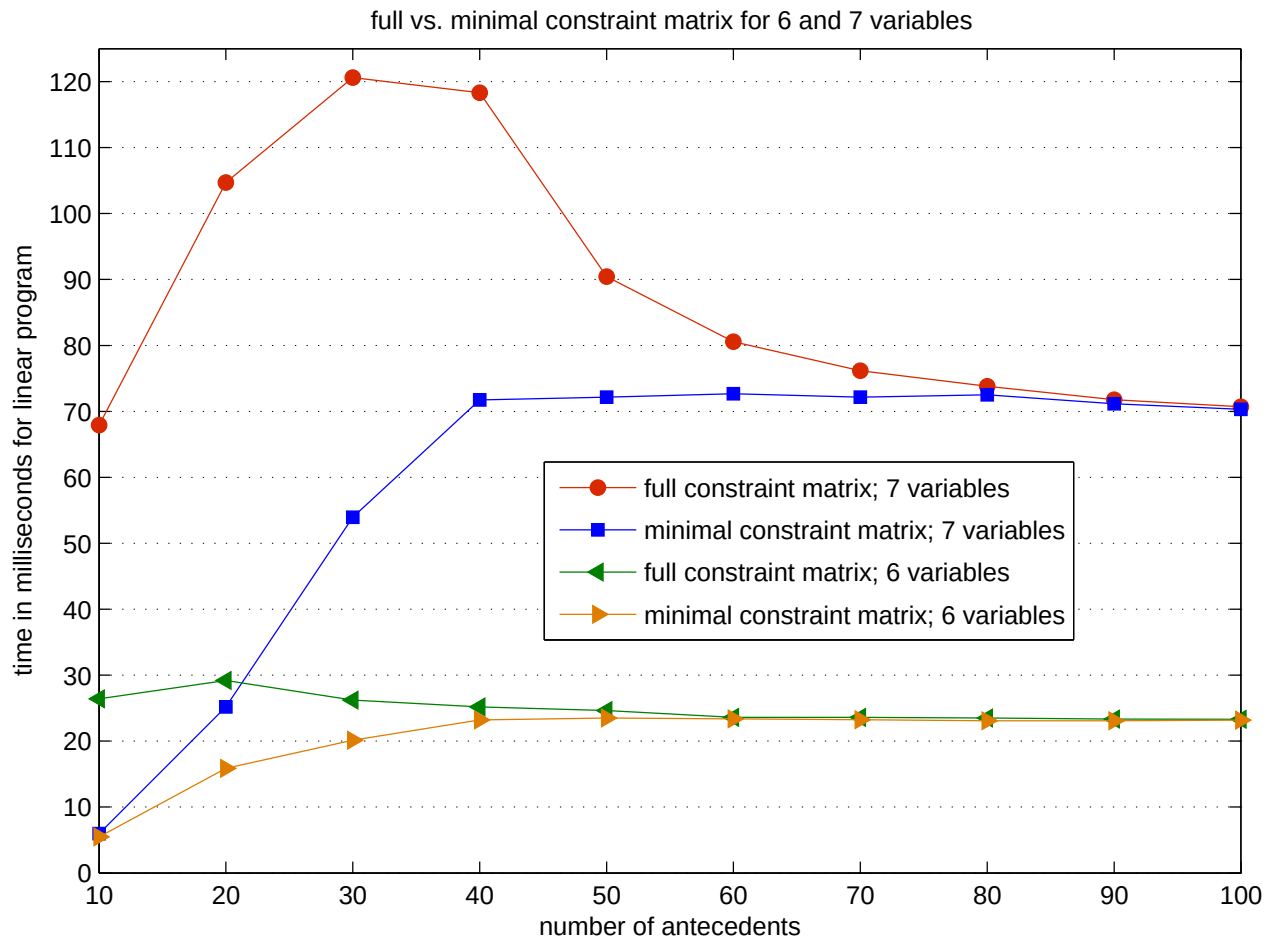

Figure 4: Average time needed [in ms; 30,000 trials] to solve a linear program with and without optimizing the constraint matrix $\mathbf{A}$; average of 30,000 trails for six and seven variables, respectively.

Figure 3 depicts the average size (that is, the number of columns and rows) of the constraint matrix $\mathbf{A}$, and the average time in milliseconds to solve one linear program and the corresponding integer program, respectively. After solving the linear programs, that is, determining whether or not there exists a solution, we also solved the equivalent integer programs. Interestingly, for each and every of the 2,700,000 instances for five variables, if there existed a solution to the linear program, then there also existed one for the corresponding integer program.

Figure 4 illustrates the computational advantage one gains when using the *minimal* constraint matrix $\mathbf{A}$ consisting only of the *relevant* elementary CI statements, over using the matrix consisting of all elementary CI statements. The times in milliseconds provided are for 6 and 7 variables, averaged over 30,000 trials, for 1000 sets of $\ell = 10, 20, ..., 100$ antecedents, and 30 different consequences, one at a time. Figure 5 depicts the average time in seconds to solve instances of the linear programs and the average dimensions of constraint matrix $\mathbf{A}$ for different number of variables, averaged over 1000 trails.

| vars | time [s] | rows(**A**) | columns(**A**) |
|---|---|---|---|
| 6 | 0.024 | 57 (57) | 239 (240) |
| 7 | 0.073 | 117 (120) | 592 (672) |
| 8 | 0.642 | 230 (247) | 1193 (1792) |
| 9 | 1.580 | 423 (502) | 1852 (4608) |
| 10 | 2.647 | 687 (1013) | 2422 (11520) |
| 11 | 7.316 | 1221 (2036) | 3699 (28160) |
| 12 | 9.038 | 2039 (4083) | 4786 (67582) |
| 13 | 20.267 | 3331 (8178) | 6863 (159744) |
| 14 | 35.969 | 4986 (16369) | 8298 (372736) |
| 15 | 91.237 | 6713 (32752) | 11024 (860160) |

Figure 5: The values are the average time [in seconds] needed to solve the linear program, and the average number of rows and columns of the constraint matrix $\mathbf{A}$; out of 1000 trials with 50 antecedents. The values in parentheses are the maximal possible values.

Finally, we want to compare our algorithm to the *racing algorithm* introduced in (Bouckaert and Studený [9]). The falsification procedure of the racing algorithm is rooted in the theory of imsets: an instance of the implication problem is falsified if one of the supermodular functions constructed by the algorithm is a counter-model for the instance of the implication problem [9]. It is heavily randomized, has super-exponential running time, and could therefore only be tested for up to 6 variables. Furthermore, the racing algorithm might falsify implications that actually do hold. This is a consequence of the fact that the class of multiinformation functions induced by discrete probability measures is a strict subset of the class of all supermodular functions. (See Examples 4.1 and 6.2 in Studený's monograph [17].) The *falsification algorithm* based on Proposition 4.2, on the other hand, ensures that falsified instance of the implication problem are guaranteed not to be valid. The validation procedure of the racing algorithm tests whether an imset that *encodes* an instance of the implication problem is combinatorial. It makes use of some ad-hoc heuristics to speed-up the computations. The validation algorithm presented here introduces two novel ideas: (1) the representation of instances of the implication problem as linear programs; and (2) the notion and construction of *minimal* constraint matrices that increase the efficiency of the algorithm.

## 6 Conclusion

Logical inference algorithms for probabilistic conditional independence statements have several important applications from checking consistency during knowledge elicitation to constraint-based structure learning of graphical models (Gandhi et al. [7]). We proved that the implication problem for CI statements is decidable, given that the size of the domains of the random variables is known and fixed. We then presented an approximate inference algorithm which combines a falsification and a novel validation algorithm. The validation algorithm represents each set of CI statements as a minimal sparse 0-1 matrix $\mathbf{A}$ and validates instances of the implication problem by solving linear programs with $\mathbf{A}$ as the constraint matrix. We demonstrated experimentally that the approximate inference algorithm is both effective and efficient in validating and falsifying instances of the implication problem. We hope that the inference algorithm will prove useful to both researchers and practitioners.

## Acknowledgments

The author wants to thank Dirk Van Gucht and Marc Gyssens for helpful discussions, and the anonymous reviewers for their valuable comments and suggestions.

## References

[1] S. Renooij, H. J. M. Tabachneck-Schijf, S. M. Mahoney (eds.) *Proceedings of the 6th UAI Bayesian Modelling Applications Workshop*, 2008.

[2] M. Druzdzel and L. C. van der Gaag. Elicitation of Probabilities for Belief Networks: Combining Qualitative and Quantitative Information. *Proceedings of the 11th Conference on Uncertainty in Artificial Intelligence*, pages 141–148, 1995.

[3] R. Dechter. Constraint Processing. Morgan Kaufmann Publishers Inc., 2003.

[4] D. Geiger and J. Pearl. Logical and algorithmic properties of conditional independence and graphical models. *The Annals of Statistics*, 21(4):2001–2021, 1993.

[5] M. Studený. Complexity of structural models. *Proceedings of the 13th Prague Conference on Information Theory, Statistical Decision Functions and Random Processes*, pages 521–528, 1998.

[6] P. de Waal and L. C. van der Gaag. Stable independence and complexity of representation. *Proceedings of the 20th Conference on Uncertainty in Artificial Intelligence*, pages 112–119, 2004.

[7] P. Gandhi, F. Bromberg, and D. Margaritis. Learning Markov Network Structure using Few Independence Tests. *Proceedings of the SIAM Conference on Data Mining*, pages 680–691, 2008.

[8] A. Tarski. A Decision Method for Elementary Algebra and Geometry, 2nd edition. University of California Press, 1951.

[9] R. R. Bouckaert and M. Studený. Racing algorithms for conditional independence inference. *Int. J. Approx. Reasoning*, 45(2):386–401, 2007.

[10] G. E. Collins. Quantifier Elimination by Cylindrical Algebraic Decomposition–Twenty Years of Progress. *Quantifier Elimination and Cylindrical Algebraic Decomposition* Springer-Verlag, pages 8–23, 1998.

[11] A. P. Dawid. Conditional Independence in Statistical Theory. *Journal of the Royal Statistical Society*, 41:1–31, 1979.

[12] M. Niepert, D. Van Gucht, and M. Gyssens. On the conditional independence implication problem: A lattice-theoretic approach. *Proceedings of the 24th Conference on Uncertainty in Artificial Intelligence*, pages 435–443, 2008.

[13] M. Niepert and D. Van Gucht. Logical Properties of Stable Conditional Independence. *Proceedings of the 4th European Workshop on Probabilistic Graphical Models*, pages 225–232, 2008.

[14] B. Sayrafi and D. Van Gucht. Differential constraints. *Proceedings of the ACM PODS Conference*, pages 348–357, 2005.

[15] F. Matúš. Ascending and descending conditional independence relations. *Transactions of the 11th Prague Conference on Information Theory*, pages 189–200, 1992.

[16] J. Pearl. *Probabilistic reasoning in intelligent systems: networks of plausible inference.* Morgan Kaufmann Publishers Inc., 1988.

[17] M. Studený. *Probabilistic Conditional Independence Structures.* Springer-Verlag, 2005.

[18] A. Schrijver. *Theory of Linear and Integer Programming.* John Wiley & Sons, Inc, 1986.

[19] B. Fitelson. A Decision Procedure for Probability Calculus with Applications. *The Review of Symbolic Logic*, 1:111-125, 2008.

[20] R. Hemmecke, J. Morton, A. Shiu, and B. Sturmfels. Three counter-examples on semigraphoids. *Combinatorics, Probability and Computing*, 17(2):239–257, 2008.